\documentclass{article}

\usepackage{microtype}
\usepackage{graphicx}
\usepackage{subcaption}
\usepackage{booktabs} 
\usepackage{multirow}

\usepackage{hyperref}

\usepackage[preprint]{icml2026}

\usepackage{amsmath}
\usepackage{amssymb}
\usepackage{mathtools}
\usepackage{amsthm}
\usepackage{tcolorbox}
\usepackage[utf8]{inputenc}
\usepackage{xcolor}
\usepackage{tcolorbox}

\definecolor{promptbg}{RGB}{235, 238, 250}

\usepackage[capitalize,noabbrev]{cleveref}

\theoremstyle{plain}

\theoremstyle{definition}

\theoremstyle{remark}

\usepackage[textsize=tiny]{todonotes}

\icmltitlerunning{Bridging On-Policy Learning and Off-Policy Knowledge through RePO}

\begin{document}

\twocolumn[
  \icmltitle{RePO: Bridging On-Policy Learning and Off-Policy Knowledge through Rephrasing Policy Optimization}



  \icmlsetsymbol{equal}{*}

  \begin{icmlauthorlist}
    \icmlauthor{Linxuan Xia}{cad}
    \icmlauthor{Xiaolong Yang}{tencent}
    \icmlauthor{Yongyuan Chen}{cad}
    \icmlauthor{Enyue Zhao}{cad}
    \icmlauthor{Deng Cai}{cad}
    \icmlauthor{Yasheng Wang}{tencent}
    \icmlauthor{Boxi Wu}{software}
  \end{icmlauthorlist}

  \icmlaffiliation{cad}{State Key Laboratory of CAD\&CG, the College of Computer Science and Technology, Zhejiang University.}
  \icmlaffiliation{tencent}{FiT, Tencent, Shenzhen, China.}
  \icmlaffiliation{software}{School of Software Technology, Zhejiang University}

  \icmlcorrespondingauthor{Boxi Wu}{wuboxi@zju.edu.cn}

  \icmlkeywords{Reinforcement Learning, Policy Optimization, Large Language Models}

  \vskip 0.3in
]



\printAffiliationsAndNotice{This work was done during an internship at Tencent.}  

\begin{abstract}
  Aligning large language models (LLMs) on domain-specific data remains a fundamental challenge.
  Supervised fine-tuning (SFT) offers a straightforward way to inject domain knowledge but often degrades the model’s generality.
  In contrast, on-policy reinforcement learning (RL) preserves generality but fails to effectively assimilate hard samples that exceed the model’s current reasoning level. 
  Recent off-policy RL attempts improve hard sample utilization, yet they suffer from severe training instability due to the forced distribution shift toward off-policy knowledge. 
  To reconcile effective off-policy knowledge absorption with the stability of on-policy RL, we propose Rephrasing Policy Optimization (RePO). In RePO, the policy model is prompted to first comprehend off-policy knowledge and then rephrase it into trajectories that conform to its own stylistic and parametric distribution. RePO dynamically replaces low-reward rollouts with these rephrased, high-quality trajectories. This strategy guides the model toward correct reasoning paths while strictly preserving on-policy training dynamics. Experiments on several benchmarks demonstrate that RePO improves hard-sample utilization and outperforms existing baselines, achieving state-of-the-art performance.
\end{abstract}

\begin{figure}[t]
    \centering
    \includegraphics[width=0.48\textwidth]{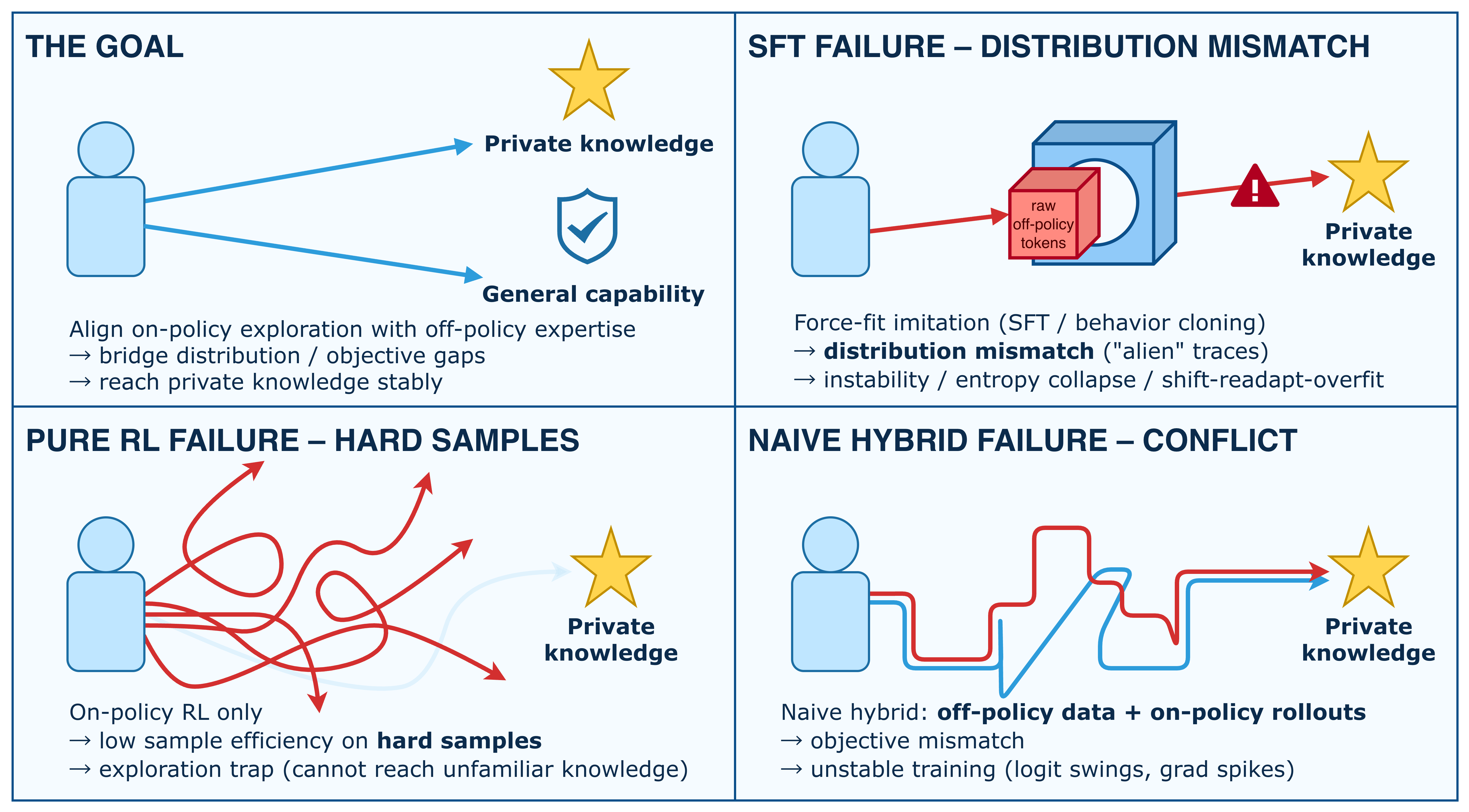}
    \caption{Challenges in bridging the gap between on-policy exploration and off-policy expertise. Aligning models with domain knowledge while maintaining general reasoning remains a fundamental challenge. SFT forces the model to fit "alien" distributions, harming its generality. Conversely, Pure RL struggles to reach unfamiliar knowledge due to the lack of guidance. Furthermore, a Naive Hybrid approach that simply mixes off-policy data results in optimization conflicts and unstable training dynamics. }
    \label{fig:teaser}
\end{figure}

\section{Introduction}

Aligning large language models (LLMs)~\cite{deepseek_r1,openai_o1} on specialized vertical domains, such as mathematics, medicine, and law, presents a fundamental alignment challenge. The objective is to inject precise domain knowledge while meticulously preserving the model's inherent general reasoning and instruction-following capabilities. A prevalent approach, supervised fine-tuning (SFT)~\cite{NEURIPS2023_ac662d74, ding2023enhancingchatlanguagemodels}, offers a straightforward mechanism for knowledge transfer. However, by learning to mimic a static dataset of expert demonstrations, SFT essentially performs off-policy training~\cite{chu2025sft}. This often leads to catastrophic forgetting of general skills and a marked degradation in the model's robust reasoning abilities, as the model overfits to the narrow distribution of the provided examples.

As an alternative, on-policy reinforcement learning (RL) has emerged as a promising paradigm for preserving model generality during fine-tuning. Techniques like Group Relative Policy Optimization (GRPO)~\cite{grpo} have been successfully applied in reasoning tasks. By generating and learning from its own outputs, the policy model undergoes updates that are constrained within its own learned distribution, thus maintaining stability and previously acquired skills. However, a critical limitation surfaces when confronting "hard samples"—queries whose solution complexity lies just beyond the model's current reasoning frontier. The on-policy paradigm, reliant on the model's own rollouts, struggles to generate correct trajectories for these samples. Consequently, it fails to effectively assimilate this high-value knowledge, creating a performance plateau.

Recent research~\cite{chord_paper}, such as LUFFY~\cite{luffy_paper}, has explored off-policy RL to overcome this knowledge absorption barrier. By leveraging off-policy knowledge from stronger models or human experts, these methods directly expose the learner to correct solutions for hard problems. Theoretically, the reuse of high-value samples can boost learning efficiency. However, in practice, forcing the current policy to fit an external, potentially dissimilar expert distribution induces a severe distributional shift.  This leads to significant training instability. The core dilemma remains unresolved: how to leverage valuable off-policy knowledge without sacrificing the training stability inherent to on-policy learning.


To bridge this gap, we propose \textbf{Rephrasing Policy Optimization (RePO)}, a novel framework designed to reconcile effective off-policy knowledge utilization with on-policy training stability. RePO introduces a two-stage knowledge assimilation mechanism. First, the policy model is prompted to comprehend an off-policy trajectory—an expert solution to a hard problem. Subsequently, it is tasked with rephrasing this solution into its own stylistic and parametric "voice." This process transforms raw expert knowledge into a native, on-policy-compatible trajectory. During the RL training loop, RePO dynamically monitors the quality of model-generated rollouts. Those falling below a reward threshold are selectively replaced by these pre-assimilated, high-quality rephrased trajectories. This strategy ensures that the learning signal for hard samples consistently guides the model toward the correct reasoning path, while the training dynamics strictly adhere to the model's own parameter distribution. Therefore, RePO  effectively combines the strengths of both on-policy and off-policy approaches.


Extensive experiments across mathematical reasoning and general knowledge benchmarks validate the efficacy of RePO. Our results demonstrate that RePO significantly outperforms both standard on-policy RL baselines and existing off-policy methods. It achieves superior utilization of hard samples, leading to state-of-the-art performance on standard evaluation sets. This work not only advances the methodology for vertical domain adaptation but also provides a principled approach to integrating heterogeneous knowledge sources in RL for language models.

\begin{figure*}[t]
    \centering
    \includegraphics[width=0.95\textwidth]{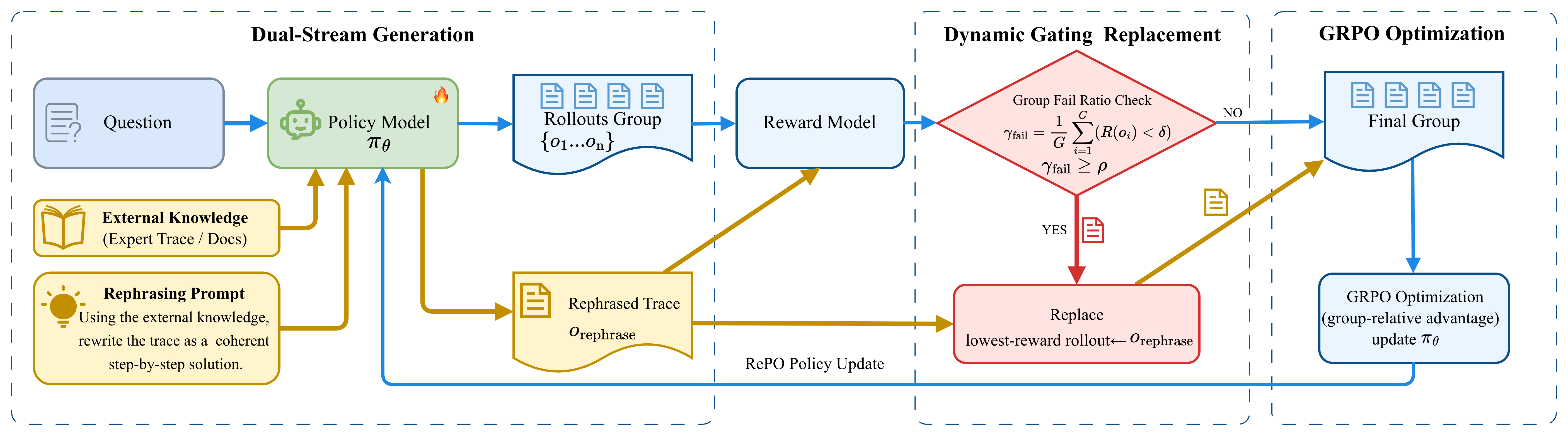}
    \caption{Overview of the Rephrasing Policy Optimization (RePO) framework. The pipeline consists of three key phases: (1) Knowledge Internalization: A rephrasing prompt guides the policy model to comprehend external knowledge and rewrite it into its native stylistic distribution, converting off-policy data into on-policy-compatible traces. (2) Dynamic Injection: To minimize instability caused by distribution shifts, the rephrased trace selectively replaces a low-quality rollout only when the group exhibits a high failure rate. (3) Optimization: The final rollout group, potentially containing the guided trace, is updated via the standard GRPO process. }
    \label{fig:pipeline}
\end{figure*}

\section{Preliminaries}

In this section, we establish the mathematical foundation for our framework. We first formalize the problem of fine-tuning LLMs using Reinforcement Learning with Verifiable Rewards (RLVR). Subsequently, we review GRPO, a representative algorithm in this domain that serves as the backbone of our study.

\subsection{Reinforcement Learning with Verifiable Rewards (RLVR)}

The process of enhancing the reasoning capabilities of LLMs can be formulated as a token-level Markov Decision Process (MDP) \cite{mdp_citation}. In this setting, the Large Language Model functions as a stochastic policy $\pi_\theta$, parameterized by $\theta$, which maps states to a probability distribution over the vocabulary.
Formally, a simplified token-level MDP tuple is $(\mathcal{S}, \mathcal{A}, \pi, \mathcal{R})$:

\textbf{State ($\mathcal{S}$):} At any time step $t$, the state $s_t$ is represented by the concatenation of the original input query $q$ and the sequence of tokens generated up to that point, denoted as $s_t = (q, o_1, o_2, \dots, o_{t-1})$. The initial state $s_0$ corresponds to the query $q$ itself.

\textbf{Action ($\mathcal{A}$):} The action space corresponds to the model's vocabulary $\mathcal{V}$. An action $a_t$ involves selecting the next token $o_t \in \mathcal{V}$.

\textbf{Policy ($\pi$):} The policy $\pi_\theta(o_t | s_t)$ defines the probability of generating token $o_t$ given the current context $s_t$. A complete trajectory (or response) $\tau = (o_1, \dots, o_T)$ is generated by sequentially sampling from this policy.

\textbf{Reward ($\mathcal{R}$):} Unlike standard RL tasks with dense signals, RLVR typically relies on a sparse, binary reward signal determined by a verifier function at the end of the generation. Let $y$ be the ground truth answer for query $q$. The reward $R(\tau)$ is assigned as:
\begin{equation}
    R(\tau) = \mathbb{1}(\text{verify}(\tau, y)) = 
    \begin{cases} 
    1 & \text{if correct}, \\
    0 & \text{otherwise}.
    \end{cases}
\end{equation}

The overarching objective of RLVR is to optimize the policy parameters $\theta$ to maximize the expected return over the distribution of queries $\mathcal{D}$:
\begin{equation}
    \mathcal{J}(\theta) = \mathbb{E}_{q \sim \mathcal{D}, \tau \sim \pi_\theta(\cdot|q)} [R(\tau)].
\end{equation}
Standard policy gradient methods~\cite{reinforce_citation} are often employed to estimate the gradients for this objective, though they face challenges related to high variance and sample inefficiency in sparse reward settings.

\subsection{Group Relative Policy Optimization (GRPO)}

GRPO has emerged as a highly effective method for reasoning tasks, particularly in mathematics and coding. A distinguishing feature of GRPO is its elimination of the value function critic, which is computationally expensive to train. Instead, GRPO estimates the baseline for the advantage function using group statistics derived from multiple sampled outputs for the same query.

Specifically, for each query $q$, the algorithm samples a group of $G$ outputs $\{o_i\}_{i=1}^G$ from the old policy $\pi_{\theta_{\text{old}}}$. By leveraging the set of rewards $\mathcal{G} = \{R(o_1), \dots, R(o_G)\}$ associated with these outputs, GRPO estimates the advantage $A_{i,t}$ for each token in output $o_i$ as:
\begin{equation}
    A_{i,t} = \frac{R(o_i) - \operatorname{mean}(\mathcal{G})}{\operatorname{std}(\mathcal{G}) + \epsilon},
\label{adv}
\end{equation}
where $\operatorname{mean}(\mathcal{G})$ and $\operatorname{std}(\mathcal{G})$ denote the mean and standard deviation of the reward group, respectively, and $\epsilon$ is a small constant for numerical stability. Notably, this advantage value is applied uniformly across all tokens in the sequence $o_i$.
The optimization objective of GRPO adopts the clipped surrogate loss characteristic of Proximal Policy Optimization (PPO) \cite{ppo_citation} to ensure training stability. To accommodate the group-based sampling, the objective function is defined as:
\begin{equation}
\begin{split}
    \mathcal{J}_{\text{GRPO}}(\theta) = \mathbb{E}_{q \sim \mathcal{D}, \{o_i\} \sim \pi_{\theta_{\text{old}}}} \Bigg[ \frac{1}{G} \sum_{i=1}^G \frac{1}{|o_i|} \sum_{t=1}^{|o_i|} \bigg( \\
    \min \left( r_{i,t} A_{i,t}, \text{clip}(r_{i,t}, 1-\epsilon, 1+\epsilon) A_{i,t} \right) \\
    - \beta \mathbb{D}_{\text{KL}}(\pi_\theta || \pi_{\text{ref}}) \bigg) \Bigg],
\end{split}
\label{grpo_obj}
\end{equation}
where $r_{i,t} = \pi_\theta(o_{i,t} | q, o_{i,<t}) / \pi_{\theta_{\text{old}}}(o_{i,t} | q, o_{i,<t})$ represents the importance sampling ratio between the current and old policies. The term $\mathbb{D}_{\text{KL}}$ represents the KL-divergence penalty used to constrain the policy update relative to a reference model $\pi_{\text{ref}}$. However, in many practical implementations of GRPO, the KL coefficient $\beta$ is often set to 0, simplifying the optimization process \cite{grpo}.

\section{Method}

\subsection{Rephrasing Policy Optimization (RePO)}

To effectively bridge the gap between high-quality off-policy guidance and the model's intrinsic exploration capabilities, we propose Rephrasing Policy Optimization (RePO). This framework is designed to enable the policy to "digest" external knowledge before assimilation, ensuring that off-policy knowledge is injected while maintaining the model's own distribution as much as possible. The methodology unfolds in two sequential phases: \textit{Joint Probability Trajectory Sampling based on Off-policy Knowledge}, which generates candidate solutions by conditioning on expert information, and a \textit{Dynamic Guidance Strategy based on Group Reward Distribution}, which selectively incorporates these solutions into the training set based on the difficulty of the query.

\paragraph{Joint Probability Trajectory Sampling based on Off-policy Knowledge.}
Standard RLVR methods sample trajectories solely based on the current policy $\pi_\theta(\cdot|q)$. In contrast, RePO leverages external off-policy knowledge to guide the generation process. Let $k$ denote the off-policy knowledge (e.g., a ground-truth reasoning path or an expert hint) associated with query $q$. We modify the sampling paradigm to a conditional probability distribution $\pi_\theta(\cdot | q, k)$, encouraging the model to reason under the guidance of $k$.
Specifically, we construct a rephrasing prompt $\mathcal{P}(q, k)$ that instructs the model to comprehend the logic within $k$ and rewrite it in its own style. By sampling from the policy given this prompt, we obtain a rephrased rollout, denoted as:
\begin{equation}
    o_{\text{rep}} \sim \pi_\theta(\cdot | \mathcal{P}(q, k)).
\end{equation}
In the context of group-based training, for each query $q$, we first sample a standard group of $G$ on-policy rollouts $\mathcal{O} = \{o_1, o_2, \dots, o_G\}$ where $o_i \sim \pi_\theta(\cdot|q)$. With the addition of the rephrased rollout $o_{\text{rep}}$, the augmented candidate pool temporarily expands to a size of $G+1$. Before reward calculation, $o_{\text{rep}}$ undergoes a post-processing step (e.g., removing prompt-specific meta-tokens) to ensure its format is identical to the standard rollouts in $\mathcal{O}$. 

\paragraph{Dynamic Guidance Strategy based on Group Reward Distribution.}
The distribution of rewards within the on-policy group $\mathcal{O}$ serves as a proxy for estimating the difficulty of the current query $q$ relative to the model's capabilities. We utilize this distribution to implement a dynamic gating mechanism that determines whether to substitute an on-policy rollout with the off-policy guided $o_{\text{rep}}$.
We introduce two hyperparameters: a reward threshold $\delta$ and a minimum failure rate $\rho$. A rollout $o_i$ is classified as a failure if its reward $R(o_i) < \delta$. Consequently, we define the group failure rate $\gamma_{\text{fail}}$ as the proportion of failed trajectories within the standard group:
\begin{equation}
    \gamma_{\text{fail}} = \frac{1}{G} \sum_{i=1}^G \mathbb{1}(R(o_i) < \delta).
\end{equation}
The replacement strategy follows a threshold-based logic. If $\gamma_{\text{fail}} \ge \rho$, it indicates that the model struggles to solve $q$ through independent exploration. In this scenario, we replace the rollout with the lowest reward in $\mathcal{O}$ with the rephrased trajectory $o_{\text{rep}}$ to provide a high-quality learning signal. Conversely, if $\gamma_{\text{fail}} < \rho$, the model is considered to have sufficient exploration capability for this query, and the original group $\mathcal{O}$ is retained to preserve on-policy distribution purity. Formally, the final training group $\mathcal{O}_{\text{final}}$ is updated as:
\begin{equation}
    \mathcal{O}_{\text{final}} = 
    \begin{cases} 
    (\mathcal{O} \setminus \{o_{\min}\}) \cup \{o_{\text{rep}}\} & \text{if } \gamma_{\text{fail}} \ge \rho, \\
    \mathcal{O} & \text{otherwise},
    \end{cases}
\end{equation}
where $o_{\min} = \arg\min_{o \in \mathcal{O}} R(o)$. Finally, the policy $\pi_\theta$ is optimized using the GRPO objective $\mathcal{J}_{\text{GRPO}}(\theta)$ (Eq. \ref{grpo_obj}) computed over the trajectories in $\mathcal{O}_{\text{final}}$.

\subsection{Training Stability Analysis}
\label{sec:stability_analysis}
RL algorithms, particularly those applied to large-scale language models, often face significant challenges regarding training stability. Instability can manifest as erratic performance, divergence of training metrics, or even catastrophic collapse, hindering the model's ability to learn effective policies. Understanding and mitigating these instabilities is crucial for successful application of RL in complex environments. This section analyzes the training stability of three distinct policy optimization methods: GRPO, LUFFY, and RePO. We investigate their inherent characteristics and their impact on key stability metrics, providing a mathematical and empirical perspective on their performance.

\subsubsection{Qualitative Comparison of Policy Optimization Methods}
We focus on three methods that aim to improve upon standard policy gradient approaches, particularly in the context of large-scale language models. Their core differences lie in how they sample data, estimate advantages, and handle off-policy corrections in Table \ref{tab:method_comparison}.

GRPO employs group-wise sampling and computes advantages relative to other rollouts within the same group. While designed to enhance stability by normalizing advantage estimates, it can suffer from low utilization when all rollouts within a group yield similar rewards (e.g., all 0s or all 1s), leading to a lack of effective update directions.

LUFFY attempts to provide more effective update directions by directly incorporating offline samples from a more advanced, high-level model. However, a critical issue arises from the vocabulary mismatch between the high-level model (source of offline samples) and the training model. To use these samples, LUFFY forcibly maps the high-level model's tokens to the training model's local vocabulary. This forced fitting generates inconsistent results during parameter update calculations. Since these inconsistent results inevitably guide parameter updates, LUFFY often leads to training instability, particularly for hard samples.

RePO addresses the limitations of LUFFY by adopting a different strategy for utilizing offline samples. Instead of direct mapping, RePO uses the training model itself to "rephrase" the offline samples. It regenerates reasonable thinking and response sequences using the training model's native vocabulary. This ensures that the parameter updates are computed based on consistent and accurate information, leading to correct model parameter updates and enhanced stability.

\begin{table}[t]
    \centering
    \caption{Comparison of Policy Optimization Methods: GRPO, LUFFY and RePO.}
    \label{tab:method_comparison} 
    \resizebox{\linewidth}{!}{
    \begin{tabular}{l | ccc}
        \toprule
        \textbf{Method} & \textbf{Sampling Strategy} & \textbf{Offline Sample Usage} & \textbf{Key Issue} \\
        \midrule
        \multirow{2}{*}{\textbf{GRPO}} & \multirow{2}{*}{Group-wise sampling} & \multirow{2}{*}{N/A} & Low utilization\\
        & & &(flat rewards) \\ 
        \midrule

        \multirow{2}{*}{\textbf{LUFFY}} & Group-wise & Direct replacement & Vocabulary mismatch \\
        &+ Offline samples &(unknown forced  vocab fit) &$\rightarrow$ instability \\ 
        \midrule

        \multirow{2}{*}{\textbf{RePO}} & Group-wise  & Training model rephrases & Consistent updates \\
        &+ Offline samples(rephrased) &(local vocab) &$\rightarrow$ stability \\ 
        \bottomrule
    \end{tabular}
    }
\end{table}

\subsubsection{Quantitative Experimental Indicators}
To quantitatively assess the training stability of these methods, we primarily monitor three key metrics: Entropy, Gradient Norm (GradNorm), and Reward. Reward is a post-training evaluation metric, reflecting the model's overall performance and ability to generate sensible responses. Entropy and GradNorm, however, are critical in-training indicators that provide insights into the policy's exploration behavior and the health of the optimization process, respectively.
\paragraph{GRPO: Stable but Low Utilization}
GRPO's stability stems from its use of group-wise sampling and relative advantage estimation.
The policy gradient update typically involves an advantage function $A(s,a)$ and the policy's log-probability $\log \pi(a|s)$:
\begin{equation} 
  \nabla \mathcal{J} = \mathbb{E}_{\tau \sim \pi} \left[ \sum_{t=0}^{T-1} \nabla \log \pi(a_t|s_t) A(s_t, a_t) \right] 
\end{equation}
In GRPO, the advantage is computed relative to other rollouts within the same group. Let $R(o_i)$ be the return of the rollout $o_i$ in a group of $G$ rollouts. The relative advantage for rollout $i$ might be formulated as Eq. \ref{adv}. This normalization ensures that updates are based on relative performance rather than absolute values, which can be more robust to noisy reward signals and prevent excessively large advantage estimates.

\begin{figure}[t]
    \centering
    \includegraphics[width=0.49\textwidth, height=0.4\textwidth]{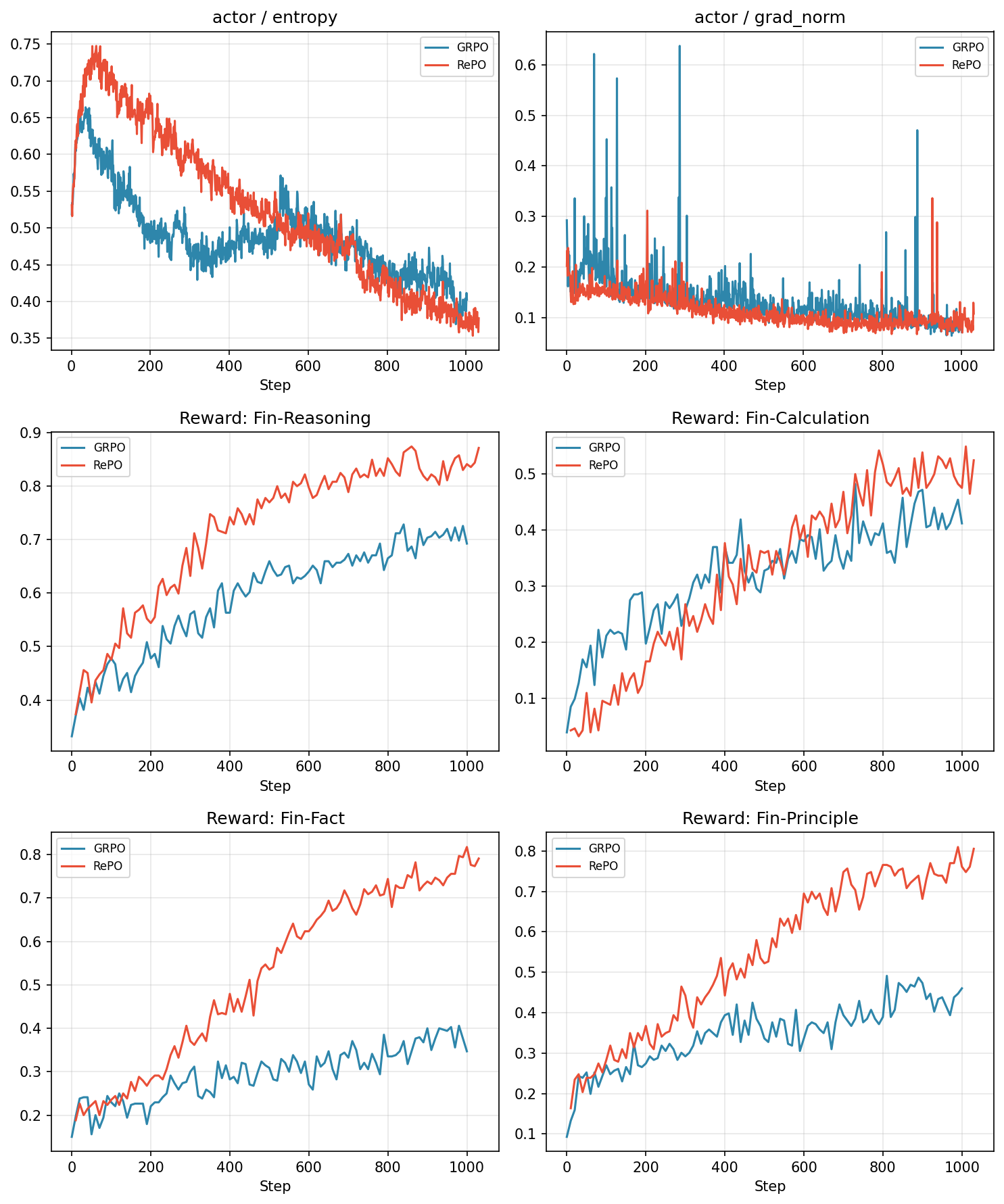}
    \caption{GRPO v.s. RePO on \textbf{Tencent FinLLM Eval}~\cite{finLLM-Eval}: Stable entropy,  stable GradNorm, and rewards with different growth rates. For reasoning tasks (reasoning and calculation, middle row), RePO shows a slight advantage over GRPO, while for knowledge-based tasks (fact and principle, bottom row), RePO's knowledge injection yields significant improvements. }
    \label{fig:grpo_stability}
\end{figure}

\begin{figure}[t]
    \centering
    \includegraphics[width=0.45\textwidth]{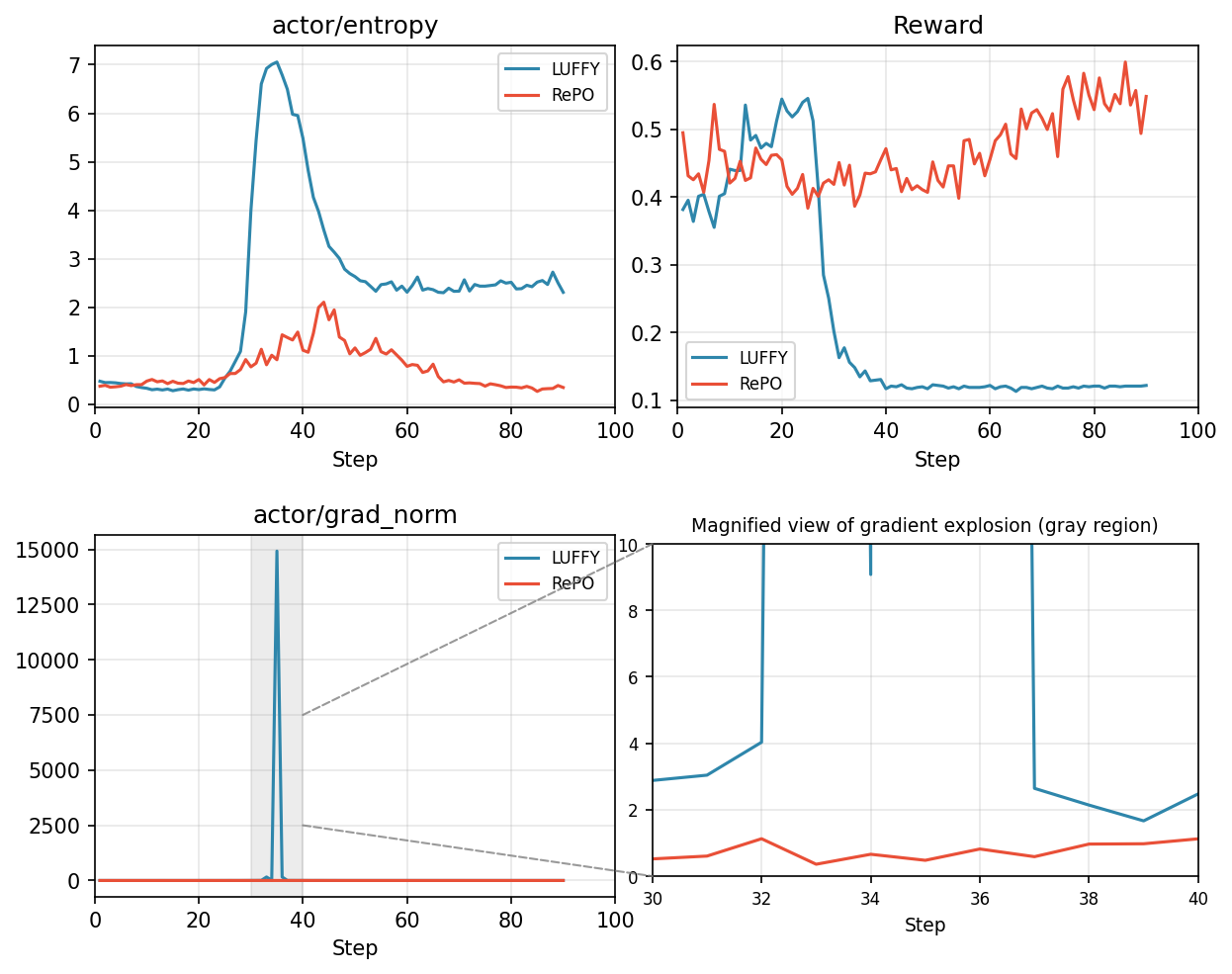}
    \caption{LUFFY v.s. RePO on the \textbf{OpenR1-Math Hard Subset}: Unstable entropy, exploding GradNorm and vanishing reward. Overly difficult samples and vocabulary mismatch rollouts during training lead to a very large and aggressive parameter update, causing gradient explosion and model collapse. }
    \label{fig:luffy_stability}
\end{figure}

\paragraph{LUFFY: Unstable due to Vocabulary Mismatch}
Its instability directly arises from its strategy of incorporating offline samples from a high-level model. The core problem lies in the vocabulary mismatch and the forced token mapping.
When offline samples are used, the policy loss depends on the log-probabilities of tokens generated by the high-level model, $\log \pi_{high}(a|s)$. If these tokens are not directly present or have different semantics in the training model's vocabulary, LUFFY attempts to map them. Let $a'_{high}$ be a token from the high-level model's vocabulary. LUFFY maps this to a token $a'_{train}$ in the training model's vocabulary. This mapping is inherently lossy and inconsistent.The policy gradient is then computed using these mapped tokens:
\begin{equation}
    \nabla \mathcal{J}_{LUFFY} = \mathbb{E}_{\text{offline}} \left[ \nabla \log \pi(a'_{train}|s) \cdot A \right] 
\end{equation}
where $A$ might still be a relative advantage.

\paragraph{RePO: Stable and Correct Updates}
RePO addresses the fundamental flaw of LUFFY by ensuring consistency between the offline samples and the training model's internal representations. Instead of directly mapping, RePO uses the training model itself to \textbf{rephrase} the offline samples into its native vocabulary. This process can be viewed as:
\begin{equation}
(s, a_{high}, R_{high}) \xrightarrow{\text{RePO}} (s, \hat{a}_{train}, \hat{R}_{train})
\end{equation}
where $\hat{a}_{train}$ is a sequence generated by the training model, representing the original $a_{high}$ but expressed using the training model's vocabulary and style. The reward $\hat{R}_{train}$ is a re-evaluated one.
The policy gradient update then becomes:
\begin{equation}
 \nabla \mathcal{J}_{RePO} = \mathbb{E}_{\text{rephrased}} \left[ \nabla \log \pi(\hat{a}_{train}|s) \cdot A \right] 
\end{equation}
Here, $\hat{a}_{train}$ is a token sequence that the training model can naturally generate and assign probabilities to.

\begin{itemize}
    \item \textbf{Logits and Entropy}: By rephrasing with its own vocabulary, RePO ensures that the target $\hat{a}_{train}$ is consistent with the training model's linguistic and semantic space.
        \begin{itemize}
            \item The logits generated for $\hat{a}_{train}$ are therefore more stable and aligned with the model's capabilities.
            \item This leads to a \textbf{stable and controlled entropy} curve. The policy can gradually adjust its distribution to assign higher probabilities to the rephrased tokens without experiencing severe internal conflicts. Entropy might show a healthy exploration pattern, initially high and gradually decreasing as the policy converges, but without erratic jumps.
        \end{itemize}
    \item \textbf{GradNorm}: The correct parameter calculation and correct model parameter updates are the direct consequence of this vocabulary alignment. Since the target $\hat{a}_{train}$ is inherently speakable by the training model, the gradients $\nabla \log \pi(\hat{a}_{train}|s)$ will be well-behaved.
        \begin{itemize}
            \item Gradients will point in directions that logically improve the policy's ability to generate desirable rephrased sequences.
            \item This prevents the generation of excessively large or conflicting gradients, resulting in a \textbf{stable and well-bounded GradNorm}. Gradient clipping can further ensure this stability, but the inherent stability of RePO's update mechanism means it is less likely to frequently hit the clipping threshold due to pathological gradients.
        \end{itemize}
    \item \textbf{Reward}: With stable entropy and well-behaved gradients, RePO facilitates consistent and effective learning. Its reward curve is expected to show steady improvement, avoiding the instabilities and collapse seen in LUFFY, and potentially achieving higher performance than GRPO due to effective utilization of offline data.
\end{itemize}

Training stability is a cornerstone for the successful deployment of deep reinforcement learning agents. Our analysis of GRPO, LUFFY, and RePO highlights how algorithmic choices, particularly concerning data sampling and off-policy correction, profoundly impact this stability. \textbf{RePO} provides a robust solution by rephrasing offline samples using the training model's native vocabulary. This ensures consistency, leading to well-behaved `logits`, stable entropy, and correctly computed gradients (stable GradNorm). Consequently, RePO achieves stable training and consistent performance improvements.

\begin{table*}[t]
\centering
\caption{Performance comparison on general knowledge and mathematical reasoning benchmarks. The models are trained on the \textbf{SuperGPQA} subset (11k non-computational problems) and evaluated on the full \textbf{GPQA} dataset for general knowledge, alongside various math benchmarks. The best results for each metric are highlighted in \textbf{bold}.}
\label{tab:main_results_knowledge}
\setlength{\tabcolsep}{5pt}
\begin{tabular}{lccccccc}
\toprule
\textbf{Method} & \textbf{GPQA} & \textbf{AIME24} & \textbf{AIME25} & \textbf{AMC} & \textbf{MATH-500} & \textbf{Minerva} & \textbf{Olympiad} \\
\midrule
Qwen3-8B & 58.1 & 75.1  & 66.4 & 88.9 & \textbf{96.2} & 51.1 & 69.2 \\
GRPO & 59.2 & 75.1 & 65.8 & \textbf{89.3} & 94.8 & 65.4 & \textbf{69.8} \\
LUFFY & 49.8 & 75.5 & 64.1 & 87.9 & 94.0 & 66.5 & 68.7 \\
\textbf{RePO (Ours)} & \textbf{61.8} & \textbf{75.8} & \textbf{72.5} & 88.6 & 94.8 & \textbf{68.1} & 68.1 \\
\bottomrule
\end{tabular}
\end{table*}

\section{Experimental Results}

In this section, we present the experimental setup, including the datasets constructed for training and evaluation, as well as the baseline models used for comparison.

\subsection{Setup}
\begin{table*}[t]
\centering
\caption{Performance comparison on the \textbf{OpenR1-Math Hard Subset} (18k samples). This subset consists of difficult problems where the reasoning chain length of DeepSeek-R1 ranges from 2k to 4k tokens. Both methods were trained with the same number of samples and evaluated on in-domain mathematical benchmarks.}
\label{tab:hard_subset}
\setlength{\tabcolsep}{8pt}
\begin{tabular}{lcccccc}
\toprule
\textbf{Method} & \textbf{AIME24} & \textbf{AIME25} & \textbf{AMC} & \textbf{MATH-500} & \textbf{Minerva} & \textbf{Olympiad} \\
\midrule
Qwen2.5-Math-7B & 11.5 & 4.9 & 31.3 & 43.6 & 7.4 & 15.6 \\
LUFFY & 0.0 & 0.0 & 0.03 & 0.0 & 0.7 & 0.0 \\
\textbf{RePO (Ours)} & \textbf{13.1} & \textbf{10.0} & \textbf{48.5} & \textbf{78.2} & \textbf{32.0} & \textbf{39.1} \\
\bottomrule
\end{tabular}
\end{table*}

\begin{table*}[t]
\centering
\caption{Performance comparison on the \textbf{OpenR1-Math Multi-source Harder Subset} (10k samples). This subset contains extremely challenging problems (response length 3k-8k tokens) distilled from multiple models (DeepSeek-R1, DeepSeek-V3.1-Terminus, DeepSeek-V3.2-Speciale, gpt-oss-120b). We evaluate both in-distribution mathematical reasoning and out-of-distribution general capabilities.}
\label{tab:multisource}
\resizebox{\textwidth}{!}{
\begin{tabular}{lccccccccc}
\toprule
\textbf{Method} & \textbf{AIME24} & \textbf{AIME25} & \textbf{AMC} & \textbf{MATH-500} & \textbf{Minerva} & \textbf{Olympiad} & \textbf{ARC-c} & \textbf{GPQA-diamond} & \textbf{MMLU-Pro} \\
\midrule
Qwen-4B & \textbf{72.7} & 65.4 & \textbf{88.2} & 96.6 & \textbf{48.9} & 67.4 & 62.4 & \textbf{54.0} & \textbf{71.3} \\
LUFFY & 61.5 & 46.4 & 81.6 & 93.8 & 47.4 & 62.5 & 33.6 & 51.0 & 70.5 \\
\textbf{RePO (Ours)} & 72.1 & \textbf{66.1} & 87.8 & \textbf{96.6} & 48.2 & \textbf{68.1} & \textbf{70.6} & 52.5 & 71.2 \\
\bottomrule
\end{tabular}
}
\end{table*}
\begin{table*}[t]
\centering
\caption{Performance comparison on Financial and Mathematical benchmarks. Both SFT and RePO methods were fully trained on the same \textbf{Financial Data} to explore the performance upper bounds on the current dataset. The metrics on financial benchmarks measure the effectiveness of domain-specific knowledge injection, while results on mathematical benchmarks indicates reasoning capabilities.}
\label{tab:financial_results}
\begin{tabular}{l|ccccc|cc}
\toprule
\multirow{2}{*}{\textbf{Method}} & \multicolumn{5}{c|}{\textbf{Knowledge Injection}} & \multicolumn{2}{c}{\textbf{Reasoning Ability}} \\
 & FinLLM-Eval & FinanceIQ & Fineva & FinEval & Finova & AIME24 & AIME25 \\ \midrule
Qwen3-8B & 30.01 & 70.3 & 73.6 & 74.6 & 55.3 & 75.1 & 66.4 \\
SFT & 39.51 & 71.7 & 74.7 & 75.1 & 57.8 & 41.7 & 28.3 \\
\textbf{RePO (Ours)} & \textbf{68.08} & \textbf{73.6} & \textbf{77.5} & \textbf{77.9} & \textbf{58.5} & 69.2 & 58.3 \\ \bottomrule
\end{tabular}%
\end{table*}

\paragraph{Datasets.} To comprehensively evaluate the performance of our method in different domains, we constructed data sets that cover both general knowledge and complex mathematical reasoning. Specifically, we curated the following datasets:

\begin{itemize}
    \item \textbf{SuperGPQA}~\cite{pteam2025supergpqascalingllmevaluation}: Designed to assess general knowledge capabilities, this dataset is a comprehensive benchmark that assesses graduate-level knowledge and reasoning capabilities across 285 disciplines. We randomly picked a subset of 11k non-computational problems, where the reasoning traces were distilled from gpt-oss-120b~\cite{openai2025gptoss120bgptoss20bmodel}. 
    
    \item \textbf{OpenR1-Math}: For mathematical reasoning tasks, we derived our training data from OpenR1-Math-45k-8192 dataset used in LUFFY~\cite{luffy_paper}. To investigate model performance under varying difficulty levels and data sources, we constructed two distinct subsets based on the length of DeepSeek-R1 responses:
    
    \textit{Hard Subset}: This subset comprises 18k challenging samples where the response length of DeepSeek-R1 falls within the range of 2k to 4k tokens. It is designed to test the model's stability on difficult problems.
    
    \textit{Multi-source Harder Subset}: To evaluate the robustness of our method on multi-source off-policy data, we curated a more challenging subset of 10k samples with response lengths ranging from 3k to 8k tokens. The reasoning trajectories in this subset were distilled from multiple advanced models, including DeepSeek-V3.1-Terminus \cite{deepseekai2024deepseekv3technicalreport}, DeepSeek-V3.2-Speciale\cite{deepseekai2025deepseekv32}, and gpt-oss-120b. 

    \item \textbf{Financial Data}: To evaluate the effectiveness of our method in domain-specific knowledge injection, we constructed a financial reasoning dataset comprising 10k samples. The reasoning trajectories were distilled from DeepSeek-V3.2-Speciale \cite{deepseekai2025deepseekv32}, retaining only correct responses to ensure data quality. As a relatively low-resource domain compared to general knowledge and mathematics, training on this dataset further examines RePO's capability of injecting domain-specific knowledge into the base model.
    
\end{itemize}



\paragraph{Evaluation.} We perform a comprehensive evaluation across both mathematical reasoning and general knowledge domains. For mathematical reasoning, we used six widely recognized benchmarks: AIME 2024, AIME 2025, AMC~\cite{numina_math_datasets}, Minerva~\cite{lewkowycz2022solvingquantitativereasoningproblems}, OlympiadBench~\cite{he2024olympiadbenchchallengingbenchmarkpromoting}, and MATH-500~\cite{hendrycksmath2021}. Due to the relatively small size of the test sets for AIME 2024, AIME 2025, and AMC, we report the \textit{avg@4} metric to ensure statistical stability, while reporting \textit{pass@1} for the remaining mathematical benchmarks. To assess general knowledge and out-of-distribution generalization, we employ ARC-c~\cite{clark2018thinksolvedquestionanswering}, GPQA-diamond, MMLU-Pro~\cite{wang2024mmluprorobustchallengingmultitask}, and the full version of the GPQA dataset~\cite{rein2024gpqa}, with all results reported as \textit{pass@1}. Specifically, the experimental setup of training on SuperGPQA and evaluating on GPQA is designed to simulate the scenario of private knowledge injection, whereas the mathematical benchmarks primarily serve to measure the model's reasoning capabilities. To evaluate the effectiveness of domain-specific knowledge injection on financial data, we employ five open-source financial benchmarks: FinanceIQ~\cite{financeIQ}, Fineva~\cite{fineva}, FinEval~\cite{FinEval}, Finova~\cite{DBLP:journals/corr/abs-2507-16802}, and FinLLM Eval~\cite{finLLM-Eval}. 
For FinanceIQ and FinLLM Eval, we conduct full-scale evaluation on the complete test sets comprising 1,349 and 629 questions, respectively. 
For remaining benchmarks, we randomly sample 1k instances each to ensure evaluation efficiency while maintaining statistical reliability. 
All financial benchmark results are reported as \textit{pass@1}. For inference, we adhere to the official recommendations for Qwen models.

\subsection{Main Results}

\paragraph{Effectiveness of External Knowledge Integration.} 
All models were trained for one single epoch in the SuperGPQA subset constructed to ensure a fair comparison of learning efficiency.
As shown in Table~\ref{tab:main_results_knowledge}, our proposed RePO demonstrates superior performance, particularly in tasks requiring the integration of external knowledge. Compared to GRPO, RePO achieves significant gains on the GPQA benchmark and the challenging AIME datasets. This indicates that while GRPO relies solely on reward signals, RePO effectively leverages the external knowledge injected to explore correct reasoning paths more efficiently, leading to better generalization on knowledge-intensive and complex reasoning tasks.
Furthermore, in contrast to LUFFY, which exhibits a notable performance drop on GPQA compared to the base model, RePO maintains robust stability. This suggests that LUFFY's method of integrating expert demonstrations might lead to instability or catastrophic forgetting of general knowledge, whereas RePO successfully assimilates off-policy data without compromising the model's foundational capabilities.

\paragraph{Stability Analysis on Hard and Multi-source Data.} 
We further investigate the stability of our method compared to LUFFY under more challenging training scenarios.
First, we analyze the model stability on the Hard Subset (Table~\ref{tab:hard_subset}). In this setting, the base model struggles with the complexity of the problems. LUFFY suffers from a catastrophic collapse. In contrast, RePO remains highly stable and achieves substantial improvements over the base model.
To simulate a realistic distillation scenario involving diverse data sources, each prompt in this dataset is associated with multiple reasoning trajectories generated by different models. During training, we randomly select one trajectory per prompt for optimization. 
As shown in Table~\ref{tab:multisource}, LUFFY exhibits a notable performance degradation compared to the base model. Conversely, RePO demonstrates superior robustness. It not only maintains competitive performance on in-distribution math tasks but also significantly improves generalization on the ARC-c benchmark. 
These results confirm that RePO is more effective and stable than LUFFY when learning from complex, multi-source off-policy data.

\paragraph{Performance on Financial Domain and Maintenance of Reasoning Ability.} In this experiment, we investigate the trade-off between domain-specific knowledge injection and the maintenance of general reasoning capabilities. 
We introduced Qwen3-30B-A3B-Instruct-2507~\cite{qwen3technicalreport} as a judge to determine whether the answers are semantically equivalent to ground truths. 
As illustrated in Table~\ref{tab:financial_results}, while SFT is capable of injecting financial knowledge with relatively low computational resource consumption, the distribution shift introduced during training leads to a catastrophic degradation in the model's reasoning abilities. In contrast, RePO leverages the advantages of the Reinforcement Learning paradigm to incorporate off-policy knowledge stably and efficiently. RePO not only achieves superior performance in knowledge injection—surpassing SFT significantly on benchmarks like FinLLM Eval—but also mitigates the impact on reasoning capabilities.

\section{Related Works}

\paragraph{RLHF.}
Reinforcement learning is widely used for LLM post-training to improve instruction following, safety, and alignment. The RLHF recipe typically combines supervised fine-tuning, preference-based reward modeling, and on-policy policy optimization (often PPO~\cite{ppo_citation}) with an explicit KL constraint to a reference model for stability~\cite{Stiennon2020,Ouyang2022}. To reduce reliance on expensive annotation, RLAIF~\cite{lee2024rlaifvsrlhfscaling} introduced model-generated preference signals.

\paragraph{RLVR.}
Recent reasoning-focused LLMs have advanced RL post-training under verifiable rewards (RLVR), where correctness is checked by deterministic rules, programs, or answer matching. To scale RLVR and reduce reliance on learned value functions, group-based sampling and advantage normalization have been adopted~\cite{grpo,deepseek_r1}; GRPO provides a PPO-like objective with group-relative baselines to lower value-model overhead and stabilize updates~\cite{grpo}. Nevertheless, purely on-policy RLVR can be sample-inefficient on low-success-rate tasks: when correct trajectories are rare, rewards become too sparse and exploration becomes the bottleneck, limiting stable improvement~\cite{luffy_paper}.

\paragraph{Mix-policy RL.}
To improve sample efficiency and robustness, recent post-training increasingly incorporates non-on-policy signals such as offline data and expert trajectories. Mix-policy optimization combines supervised signals for retention/stability with RL for goal-directed improvement. At the loss level, UFT unifies SFT and reinforcement fine-tuning to mitigate overfitting and low-success exploration~\cite{Liu2025}, while SRFT uses entropy-aware weighting to jointly learn from demonstrations and on-policy rollouts~\cite{Fu2025}. For RLVR-style reasoning, LUFFY bridges on-policy updates with off-policy reasoning traces via regularized importance-sampling-based policy shaping to balance imitation and exploration and avoid rigid copying and entropy collapse~\cite{Yan2025}. At the data level, expert traces can also guide rollout generation: BREAD inserts short expert hints and branches rollouts to densify reward signals~\cite{Zhang2025}, and Prefix-RFT blends SFT and RFT through prefix sampling with minimal pipeline changes~\cite{Huang2025}.

\section{Conclusion}

In this paper, we introduced \textbf{Rephrasing Policy Optimization (RePO)}, a novel framework designed to harmonize high-quality off-policy guidance with on-policy exploration. By employing a \textit{comprehend-and-rephrase} mechanism, RePO effectively transforms external expert traces into on-distribution learning signals, thereby overcoming the distribution mismatch inherent in SFT and the exploration inefficiency of pure RL. Extensive experiments across mathematical reasoning and financial domains demonstrate that RePO achieves state-of-the-art performance, enabling the stable injection of private domain knowledge while robustly preserving the model's general reasoning capabilities.



\bibliography{RePO}
\bibliographystyle{icml2026}

\newpage
\appendix
\onecolumn
\section{Experimental Details}
\paragraph{Data Curation.} During the distillation process utilizing \texttt{gpt-oss-120b}, we configure the temperature to $1$ and $top\_p$ to $1$ to encourage the model to freely sample trajectories. Conversely, for the DeepSeek series models, we adhere to the official recommended inference parameters.

\paragraph{RL Training Settings.} In our reinforcement learning experiments, we generate $8$ rollouts for each prompt. The training is conducted with a learning rate of $1\text{e-}6$ and a batch size of $512$. Unless otherwise specified, we use rule-based reward for training. Regarding the implementation frameworks, RePO is built upon \texttt{verl}, whereas LUFFY is implemented using its official open-source codebase\footnote{\url{https://github.com/ElliottYan/LUFFY}}.

\paragraph{Model Configurations.} For the Qwen2.5 series models, we adopt the same configuration modifications as employed in LUFFY. In contrast, the Qwen3 series models are utilized without any architectural or configuration alterations.

\paragraph{Supervised Fine-tuning (SFT).} We perform SFT on the identical financial dataset utilized for the proposed RePO. The training is configured with a learning rate of $1\text{e-}5$, a learning rate warmup ratio of $0.05$, and a maximum sequence length of $16$K tokens.

\subsection{Prompt Example}

\begin{tcolorbox}[
    colback=promptbg,    
    colframe=black,      
    boxrule=1pt,         
    sharp corners,       
    boxsep=5pt,          
    left=2pt, right=2pt, top=2pt, bottom=2pt 
]
    You will be provided with a problem along with a reference reasoning process and solution for your study. Your task is to thoroughly internalize and master the problem-solving approach from the reference material, then independently reconstruct and articulate the complete reasoning process in your own style, as if you are solving the problem from scratch based on your own expertise. Critically, your response must read as purely independent reasoning: do not mention, reference, acknowledge, or hint at having seen any external guidance, reference material, or pre-existing solution.

Question: \{question\}

Reasoning Process: \{thinking\}

Reference Answer: \{response\}

Your response must strictly follow the output format below. Do not output any extraneous content:

$<$COT$>$

This is your detailed thinking process. You need to rephrase the reasoning process above in your own words. Do not omit or simplify any part of the content.

$<$/COT$>$

$<$RESPONSE$>$

This is the final answer or conclusion. Based on the thinking process you have just articulated, answer the original question with the complete calculation or analysis process.

$<$/RESPONSE$>$
\end{tcolorbox}


\end{document}